\documentclass[11pt,a4paper]{article}

\usepackage{emnlp2021}
\usepackage{times}
\usepackage{latexsym}
\usepackage{booktabs}
\usepackage{multirow}
\usepackage{makecell}
\usepackage{pifont} 
\usepackage{tabularx}
\usepackage{xspace}
\usepackage{microtype} 
\usepackage{url}
\usepackage{graphicx}
\graphicspath{ {./} }
\usepackage{natbib}
\usepackage[utf8]{inputenc}
\usepackage[T1]{fontenc}
\usepackage{nth}
\usepackage{tikz}
\usepackage{pgfplots}
\usepgfplotslibrary{groupplots}
\usepackage{stix}
\usepackage{soul}
\usepackage{adjustbox}
\usepackage{placeins}

\expandafter\def\expandafter\UrlBreaks\expandafter{\UrlBreaks
  \do\a\do\b\do\c\do\d\do\e\do\f\do\g\do\h\do\i\do\j%
  \do\k\do\l\do\m\do\n\do\o\do\p\do\q\do\r\do\s\do\t%
  \do\u\do\v\do\w\do\x\do\y\do\z\do\A\do\B\do\C\do\D%
  \do\E\do\F\do\G\do\H\do\I\do\J\do\K\do\L\do\M\do\N%
  \do\O\do\P\do\Q\do\R\do\S\do\T\do\U\do\V\do\W\do\X%
  \do\Y\do\Z}

\usepackage{soulutf8}
\usepackage[normalem]{ulem}
\usepackage{xcolor}

\usepackage{enumitem}
\setlist{nolistsep}
\usepackage{microtype}

\iftrue 
    \usepackage[final]{proofing}
  \renewcommand\hl[1]{{#1}}  
   \newcommand\draftHL[2]{#1}{\draftnote{\red{#2}}}
   \newcommand\redHL[1]{}
  \newcommand\todo[1]{}

  \newcommand{\Djame}[1]{}

\else  

\usepackage[draft]{proofing}

\newcommand{\Djame}[1]{
\textbf{\textcolor{red}{\hl{Djame: #1}}}
}

\newcommand\red[1]{{{\textcolor{red}{\bf #1}}}}

\newcommand\draftHL[2]{\hl{#1}{\draftnote{\red{#2}}}}

\let\oldred\red
\renewcommand\red[1]{{ \oldred{{#1}}}}

 \renewcommand\draftHL[2]{\hl{#1}{\draftnote{\red{#2}}}}
 \newcommand\redHL[1]{\red{\hl{#1}}}
\let\olddraftnote\draftnote
\renewcommand\draftnote[1]{\olddraftnote{\red{#1}}}

\fi

\def\aclpaperid{***} 

 \newcommand{\crapbank}{\texttt{PFSMB}\xspace}

\newcommand{\mtnt}{\texttt MTNT\xspace}
\newcommand{\wmt}{\texttt{WMT}\xspace}

\newcommand{\opentest}{\texttt{OpenSubTest}\xspace}

\newcommand{\newstest}{\texttt{NeswTest}\xspace}
\newcommand{\opensubs}{\texttt{OpenSubs}\xspace} 
\newcommand{\seqtseq}{\texttt{Seq2seq}\xspace}
\newcommand{\transformer}{\texttt{Transformer}\xspace}
\newcommand{\chartchar}{\texttt{char2char}\xspace}

\newcommand{\blue}[1]{\textcolor{blue}{#1}}
\newcommand{\unk}{\texttt{<UNK>}\xspace}

\newcommand\review[3]{\textcolor{red}{\textbf{R#1.#2.}}\textcolor{blue}{#3}}
\iftrue
\renewcommand\review[3]{#3}
\fi

\title{Noisy UGC Translation at the Character Level: Revisiting Open-Vocabulary Capabilities and Robustness of Char-Based Models
}
\iffalse
\author{José Carlos Rosales Núñez \\
Université Paris-Saclay \& CNRS, LISN \\
  91403 Orsay, France \\
  Inria Paris \\
  \texttt{jose.rosales@limsi.fr} \\\And
    Guillaume Wisniewski \\
  LLF, Université de Paris \& CNRS \\
  F-75013 Paris, France \\
  \texttt{guillaume.wisniewski@u-paris.fr} \\\And
  Djamé Seddah \\
  Inria Paris \\
  F-75012, Paris, France \\
  \texttt{djame.seddah@inria.fr} }

\else

\author{José Carlos Rosales Núñez$^{1,2,4}$ \quad Guillaume Wisniewski$^3$ \quad Djam\'e Seddah$^4$\\
$^1$ CNRS, LISN \quad $^2$ Universit\'e Paris-Saclay \\
$^3$ Université de Paris, LLF, CNRS \quad $^4$ Inria Paris\\
{\tt jose.rosales@limsi.fr} \quad {\tt guillaume.wisniewski@univ-paris-diderot.fr} \\{\tt djame.seddah@inria.fr}
   \\}

  \fi

\date{}

\definecolor{mygrey}{RGB}{229,229,229}
\definecolor{mygrey2}{RGB}{127,127,127}
\definecolor{mygrey3}{RGB}{240,240,240}

\newcommand\gray[1]{{{\textcolor{mygrey2}{#1}}}} 
 
\pgfplotsset{
 	axis background/.style={fill=mygrey},
	tick style=mygrey2,
	tick label style=mygrey2,
	grid=both,
	xtick pos=left,
	ytick pos=left,
	tick style={
		major grid style={style=white,line width=1pt},minor grid style=mygrey3,
		tick align=outside,
	},
	minor tick num=1,
}

\begin{document}
\maketitle

\begin{abstract}

  \iffalse 
This work explores the capacities of character-based Neural
  Machine Translation to translate noisy User-Generated Content
  \draftremove{the specificities (high rate of out of vocabulary
    tokens, rare or ambiguous grammatical constructs)}, which raise
  many challenges. In order to analyze the limits of these systems, we
  have manually annotated 400 UGC sentences indicating the kind of
  errors/noise they contain, which we make available for further
  exploring the impact of such specificities.  Our results show that
  character-based neural machine translation is very sensitive to UGC
  idiosyncrasies and they are not able to translate texts that are
  productive in terms of new forms, new structures or new domains.
 
  \else This work explores the capacities of character-based Neural
  Machine Translation to translate noisy User-Generated Content (UGC) with a
  strong focus on exploring the limits of such approaches to
  handle productive UGC phenomena, which almost by definition, cannot
  be seen at training time. Within a strict zero-shot scenario, we
  first study the detrimental impact on translation performance of
  various user-generated content phenomena on a small annotated dataset we 
  developed, and then show that such models are indeed incapable
  of handling unknown letters, which leads to catastrophic translation
  failure once such characters are encountered. We further confirm
  this behavior with a simple, yet insightful, copy task experiment
  and highlight the importance of reducing the vocabulary size
  hyper-parameter to increase the robustness of character-based models
  for machine translation.

\fi

\end{abstract}


\section{Introduction \label{sec:intro}}

Neural Machine Translation (NMT) models fall far short from being able
to translate noisy User-Generated Content (UGC): the quality of their
translation is often even worse than that of a traditional
phrase-based
system~\cite{DBLP:conf/aclnmt/KhayrallahK18,rosales-nunez-etal-2019-comparison}.
In addition to ambiguous grammatical constructs and profusion of
ellipsis, the main difficulty encountered when translating UGC is the
high number of out-of-vocabulary tokens (OOVs) resulting from
misspelled words, emoticons, hashtags, mentions, and all specific
constructs used in online forums and social medias
\cite{foster:2010:cba,seddahetal:2012:crapbank,eisenstein2013bad,sanguinettietal:2020:treebankingUGC}.
Some of those phenomena can be perceived as noise while the others
are typical markers of language variation among speakers. Moreover,
a certain amount of those same phenomena operate at the lexical
level (either at the character, subword or word levels) \cite{sanguinettietal:2020:treebankingUGC}. This is why, focusing more on the noise axis, char-based models appear to offer a
natural solution to this  problem~\cite{luong16achieving,ling15character}: indeed these open vocabulary models are designed specifically to address the OOV problem.

In this work, we explore the ability of out-of-the-box character-based
NMT models \cite{DBLP:journals/tacl/LeeCH17_et_al} to address the
challenges raised by UGC translation. While character-based models may
seem promising for such task, to the best of our knowledge, they have
only been tested either on data sets in which noise has been
artificially added through sampling an edited word error data set
\cite{DBLP:conf/iclr/BelinkovB18, ebrahimi18adversarial} and on
canonical data set, in which they prove to be very effective for
translating morphologically-rich languages with a high number of OOVs 
\cite{luong16achieving}.

However, our starting-points experiments show that
character-based systems are outperformed by BPE models even when
translating noisy UGC. To understand this counter-intuitive result, we
conduct several experiments and analyses. In
particular, we manually annotated 400 sentences at the token
level using a fine-grained typology, to perform our
analyses. These sentences correspond to the
worst and the best translated utterances of two MT systems (a char-based and a
transformer-based model). Our results highlight the extreme sensibility of
character-based models to the vocabulary size, a parameter often
overlooked in the literature. Using a simple set of experiments, we thus
show that these models are unable to perform an {\em easy} copy task due to their
poor handling of unknown and rare characters. By adjusting the
vocabulary size parameter, we drastically improve the robustness of
our character-based model without causing a large drop in in-domain performance.

Our contributions are as follows:
\begin{itemize}
\item we provide an annotated data set\footnote{\url{https://github.com/josecar25/char_based_NMT-noisy_UGC}} that enables in-depth evaluations
  of the impact of UGC idiosyncrasies;
\item we demonstrate that char-based neural machine translation models are
extremely sensitive to unknown and rare characters on both synthetic 
data and noisy user-generated content;
\item we show how an overlooked hyper-parameter drastically improve char-based MT models
  robustness to natural noise while maintaining the in-domain level of
  performance.\draftnote{petit bras non ? -ds} 

\end{itemize}

\section{Character-level NMT \label{sec:rw}}

In this work, we explore two character-based translation models.  The
first model we consider, \texttt{charCNN} is a classic encoder-decoder
in which the encoder uses character-based embeddings in combination
with convolutional and highway layers to replace the standard look-up
based word representations. The model considers, as input, a stream of
words (i.e.\ it assumes the input has been tokenized beforehand) and
tries to learn a word representation that is more robust to noise by
unveiling regularities at the character level. This architecture was
initially proposed by~\newcite{DBLP:conf/aaai/KimJSR16} for language
modeling; \newcite{DBLP:conf/acl/Costa-JussaF16} shows how it can be
used in an NMT system and report improvements up to 3 \textsc{Bleu}
\cite{DBLP:conf/acl/PapineniRWZ02} points when translating from a morphologically-rich language, German,
to English.

The second model we consider does not rely on an explicit segmentation
into words: \newcite{DBLP:journals/tacl/LeeCH17_et_al} introduce the
\texttt{char2char} model that directly maps a source characters
sequence to a target characters sequence without any segmentation
thanks to a character-level convolutional network with max-pooling at
the encoder. It can be considered as an open-vocabulary model:
it can generate any word made of any of the $N$ most frequent
characters of the train set (where $N$ is a model hyper-parameter) and
only outputs an \unk token in the presence of character that is not in
this (char-) vocabulary.  \newcite{DBLP:journals/tacl/LeeCH17_et_al}
show that this model outperforms subword-level (i.e.\ BPE-based)
translation models on two WMT'15 tasks (de-en and cs-en) and gives
comparable performance on two tasks (fi-en and
ru-en). \newcite{DBLP:journals/tacl/LeeCH17_et_al} additionally
  report that in a multilingual setting, the character-level encoder
  significantly outperforms the subword-level encoder on all the
  language pairs.

These two models have been originally tested on WMT or IWSLT tasks
that consider texts that mostly qualify as canonical with very few
spelling or grammatical errors. The impact of noise on
\texttt{charCNN} and \texttt{char2char} has been evaluated by
\newcite{DBLP:conf/iclr/BelinkovB18} and
\newcite{DBLP:conf/coling/EbrahimiLD18}. By adding noise to canonical
texts (the \texttt{TEDTalk} dataset). Their results show that the
different character levels models fail to translate even moderately
noisy texts when trained on `clean' data and that it is necessary to
train a model on noisy data to make it robust. Note that, as explained
in Section~\ref{sec:data}, there is no UGC parallel corpus large
enough to train a NMT model and we must rely on the models' ability to
learn, from canonical data only, noise- and error-resistant
representations of their input that are robust to the noise and errors
found in UGCs. This is why, in this work, we are interested in studying MT performance in a zero-shot \draftreplace{fashion}{scenario} when translating noisy UGC.

To the best of our knowledge, no work has studied the performance of
the character-based NMT architectures on an actual UGC scenario with
real-world gathered and fine-grained annotated noisy sentences. This sets the main motivation for the present work.

\section{Datasets \label{sec:data}}
\paragraph*{Training sets \label{sec:train_data}} Due to the lack
of a large parallel corpus of noisy sentences, we train our systems
with `standard' parallel datasets, namely the corpora used in the
\texttt{WMT} campaign~\cite{bojar16findings} and the
\texttt{OpenSubtitles} corpus~\cite{lison18opensubtitles2018}. The
former contains canonical texts (2.2M sentences), while the latter
(9.2M sentences) is made of written dialogues from popular tv series.
The reason for using \texttt{OpenSubtitles} is the assumed 
greatest similarity, compared to \texttt{WMT}, between its content and
the UGC conversational nature that predominates most social media. However, 
UGC differs significantly from subtitles
in many aspects: in UGC, emotions are often denoted with repetitions,
there are many typographical and spelling errors, and sentences may
contain emojis that can even replace some words (e.g. $\color{red}
\varheartsuit$ can stands for the verb `love' in sentences such as `I
$\color{red} \varheartsuit$ you'). UGC productivity limits the
pertinence of domain adaptation methods such as fine-tuning, as there
will always be new forms that will not have been seen during training~\cite{DBLP:conf/aclnut/AlonsoSS16}.
\draftreplace{Hence we have not considered these techniques in this
  work.}{This is why we focus here on zero-shot scenarios, as we
  believe they provide a clearer experimental protocol when it comes
  to study the impact of UGC specificities on MT models.}

\begin{table*}[!h]
{\footnotesize
  \begin{tabular}{ccrrrrrr}
    \toprule
    Corpus & Split & \#sentences & \#tokens & avg. sent.len & TTR &
    Vocab. size  & \#chars types \\
    \midrule
    \multirow{2}{*}{\makecell{\wmt}} &  Train & 2.2M  & 64.2M  & 29.7
    &  0.20 & 1.037M & 335\\   
& NewsDiscussDev & 1,500 & 24,978 & 16.65 & 0.22 & 7,692 & \\
    \midrule
\opensubs &  Train & 9.2M  & 57.7M  &    6.73 &  0.18 & 2.08M  & 428  \\
    \midrule


    \crapbank$^\dag$ & \multirow{4}{*}{Test} & 777    & 13,680 & 17.6 &  0.32 & 3,585 & 116\\
     MTNT$^\dag$ &  & 1,022 &  20,169 & 19.7 & 0.34 & 3,467 & 122\\
 OpenSubsTest {\tiny (2018)} & & 11,000 & 66,148 & 6.01 & 0.23 & 15,063 & 111 \\
  NewsTest {\tiny (2014)} &  & 3,003 & 68,155 & 22.70 & 0.23 & 15,736 & 111\\

    \bottomrule
  \end{tabular}
  }
  \centering
  \caption{Statistics on the FR side of the corpora used in our experiments. TTR stands for Type-to-Token Ratio. Corpus
    containing mainly UGC are indicated with a
    $\dag$.\label{tab:stat_corpus}}
\end{table*}

\begin{table*}[!h]
{
{\footnotesize
  \begin{tabular}{cccccc}
    \toprule
     $\downarrow$ Metric  / Test set $\rightarrow$ & \crapbank$^\dag$ & \mtnt$^\dag$ & \newstest & \opentest  \\
    \midrule
    3-gram KL-Div & 1.563 & 0.471 & 0.406  & 0.006\\
    \%OOVs  & 12.63 & 6.78 & 3.81 & 0.76 \\
    PPL  &  599.48 & 318.24 & 288.83 & 62.06 \\
         \bottomrule
  \end{tabular}
}
}
  \centering
  \caption{Domain-related measure on the source side (FR), between
    Test sets and \opensubs training set. Dags indicate
    UGC corpora.\label{tab:stat_test}}
\end{table*}


\paragraph*{Test sets} We consider in our experiments two
French-English test sets made of user-generated content. These corpora
differ in the domain of their contents, their collection date, and in the
way sentences were filtered to ensure they are sufficiently different
from canonical data.

The first one, the MTNT corpus~\cite{michel2018mtnt}, is a
multilingual dataset that contains French sentences collected on
\texttt{Reddit} and translated into English by professional
translators. The second one, the Parallel French Social Media Bank (\crapbank)\footnote{\url{https://gitlab.inria.fr/seddah/parallel-french-social-mediabank}}, introduced in
\cite{rosales-nunez-etal-2019-comparison} \draftadd{and made from a subpart of
the French Social Media Bank \cite{seddahetal:2012:crapbank}}, consists of comments
extracted from Facebook, Twitter and Doctissimo, a health-related
French forum. Table~\ref{tab:UGC_examples} shows some examples of
source sentences and reference translations extracted from these two
corpora and illustrates the peculiarities of UGC and difficulties of
translating them. User-generated contents raise many challenges for MT
systems: they notably contain \texttt{char-OOVs}, that is to say
characters that have not been seen in the training data (e.g.\
emojis), rare character sequences (e.g. inconsistent casing or
usernames) as well as many \texttt{word-OOVs} denoting URL, mentions,
hashtags or, more generally, named entities. See
\cite{foster:2010:cba,seddahetal:2012:crapbank,eisenstein2013bad,sanguinettietal:2020:treebankingUGC}
for a complete account of UGC idiosyncrasies.
\review{2}{6 - ok pour moi -ds}{We have also selected a pair of blind UGC test sets, 
corresponding to non-overlapping partitions of  the \crapbank and \mtnt datasets, 
respectively. These were chosen to validate our approaches and models, after 
performing error analysis and benchmarks with the UGC test sets described above.}

Additionally, we also evaluate our NMT systems on a pair of non-noisy,
edited, datasets, which are chosen to match both of the train data configuration
mentioned above, namely \newstest and \opentest, for \wmt and
\opensubs corpora, respectively. These test sets serve a double
purpose: evaluate the performance impact due to domain drift and
the in-domain performance for both \wmt and \opensubs. See Tables~\ref{tab:stat_corpus} \& \ref{tab:stat_test} for relevant statistics.


It is worth noticing how the noisy UGC corpus, i.e. \crapbank and
\mtnt, have remarkably high TTR, high KL-divergence, \%OOV rate and
perplexity, even when compared to the out-of-domain test set in Table~\ref{tab:stat_test} (\newstest).

\section{NMT Models \label{sec:model}}

\paragraph*{Character-based models}

In our experiments, we compare the two character-level NMT systems
introduced in Section~\ref{sec:rw}, namely \texttt{charCNN} and
\texttt{char2char}. Both models were trained as out-of-the box systems
using the implementations provided by \newcite{DBLP:conf/aaai/KimJSR16}
for
\texttt{charCNN},\footnote{\url{https://github.com/harvardnlp/seq2seq-attn}}
and by \newcite{DBLP:journals/tacl/LeeCH17_et_al} for
\texttt{char2char}.\footnote{\url{https://github.com/nyu-dl/dl4mt-c2c}}

It must be noted that the \texttt{charCNN} extracts character n-grams
for each input word and predicts a word contained in the target
vocabulary or a special token, \unk, otherwise, whereas the
\texttt{char2char} is capable of open-vocabulary translation and does
not generate \unk tokens, unless an out-of-vocabulary character
(\texttt{char-OOV}) is present in the input.

\paragraph*{BPE-based models}

We use as our baselines two standard NMT models that consider
tokenized sentences as input. 

The first one is a \texttt{seq2seq}
bi-LSTM architecture with global attention decoding. The \texttt{seq2seq} model was trained using the XNMT
toolkit~\cite{DBLP:conf/amta/NeubigSWFMPQSAG18}.\footnote{We decided
  to use XNMT, instead of OpenNMT in our experiments in order to
  compare our results to the ones of \newcite{michel2018mtnt}.} It
consists of a 2-layered Bi-LSTM layers encoder and a 2-layered Bi-LSTM
decoder. It considers, as input, word embeddings of size 512 and each
LSTM units has 1,024 components.

Our second baseline model is  a vanilla Transformer model
\cite{DBLP:conf/nips/VaswaniSPUJGKP17} using the implementation
proposed in the OpenNMT framework
\cite{DBLP:conf/amta/KleinKDNSR18}. It consists of 6 layers with word
embeddings that are 512-dimensional, a feed-forward layers made of 2,048
units and 8 self-attention heads.

\paragraph*{Unknown Token Replacement   }

One of the peculiarities of our UGC datasets is that they contain a
many OOVs denoting URL, mentions, hashtags, or more generally
named entities: for instance several sentences of the \crapbank
mention the game ``Flappy Bird'' or the TV show ``Teen Wolf''. Most of
the time, these OOVs are exactly the same in the source and target
sentences and consequently, the source and reference share a lot of
common words: the \textsc{Bleu} score between the sources and
references of the \crapbank being of 15.1,\footnote{The sources and
  references of the MTNT corpus are less similar (\textsc{Bleu} score:
  5) since all user mentions and URL are removed in the distributed
  version of the corpus.} while it is only 2.7 on the \texttt{WMT}
test set.

Closed-vocabulary MT systems that are not able to copy OOVs from the
source sentence are therefore penalized. That is why, as part of our
translation pipeline, we introduce a post-processing step in which the
translation hypothesis is aligned with the source sentence and the
\unk tokens replaced by their corresponding aligned source tokens. For
the \texttt{seq2seq} the alignments between the source and translation
hypothesis are computed using an IBM2 model.\footnote{The alignments
  are computed by concatenating the test set to the training set and
  using \texttt{FastAlign} \cite{dyer-etal-2013-simple}.}. For the
\texttt{charCNN} model, the alignments are deduced from the attention
matrix. The \texttt{char2char} model is an open-vocabulary system that
is able to generate new words when necessary. The
\draftadd{vanilla} \texttt{Transformer} implementation we use is able to copy unknown
symbols directly.
    
\section{Experiments and Results}
\label{sec:exp}
We train and evaluate the MT models presented in the previous
  section using the data train and test configurations in
  Section~\ref{sec:data}.  We first  compare
  performances for our clean test sets (whether in-domain or
  our-of-domain, accordingly) and noisy UGC when translated using BPE-
  and character-based models \draftHL{before exploring into detail our results.}{nul -ds}

\begin{table*}[!htpb]

\centering

{\footnotesize

\begin{tabular}{lcccccccccc}
\toprule
   & \phantom{ab} & \multicolumn{4}{c}{\texttt{WMT}} & \phantom{abc} &\multicolumn{4}{c}{\texttt{OpenSubtitles}}  \\
  \cmidrule{3-6} \cmidrule{8-11}
   &  \phantom{ab} & \crapbank & MTNT & News\dag &  Open & & \crapbank & MTNT &  News &  Open\dag   \\
  \midrule
  \multicolumn{11}{l}{\textit{BPE-based models}} \\
  \phantom{ab}\texttt{seq2seq}           & \phantom{ab} & \phantom{0}9.9 & 21.8         & 27.5                & 14.7         & & 17.1 & 27.2 & 19.6 & 28.2 \\ 
  \phantom{abcd} + \unk rep. & \phantom{ab} & \textbf{17.1}  & \textbf{24.0}& \textbf{29.1} & \textbf{16.4} & & 26.1 & 28.5 & 24.5 & 28.2 \\ 
  \phantom{ab}\texttt{Transformer}\\
  \phantom{abcd} + \unk rep. & \phantom{ab} & 15.4 & 21.2 & 27.4 & \textbf{16.4} & & \textbf{27.5} & \textbf{28.3} & \textbf{26.7} & \textbf{31.4}  \\ 
  \midrule
  \multicolumn{11}{l}{\textit{Character-based models}} \\
  \phantom{ab}\texttt{charCNN}   & & \phantom{0}6.2 & 12.7 & 17.2 & \phantom{0}9.2  & & 13.3 & 16.3 & 10.1 & 21.7 \\
  \phantom{abcd} + \unk rep.     & & 16.1           & 18.2 & 22.1 &           11.5  & & 18.6 & 20.2 & 14.6 & 23.9 \\
  
  \phantom{ab}\texttt{char2char} & & \phantom{0}7.1 & 13.9 & 18.1 & \phantom{0}8.8  & & 23.8 & 25.7 &  17.8 & 26.3\\ 
  \bottomrule
\end{tabular}%
\caption{\textsc{Bleu} scores for our models for the different train-test
  combinations.  In-domain test sets are marked with a dag. 
 `News' and `Open' stand, respectively, for the \texttt{WMT} and \texttt{OpenSubtitles} test
  sets. \wmt and \texttt{OpenSubtitles} are the training corpora,
  described in Section~\ref{sec:train_data}
  \label{tab:res_bleu}}
}

\end{table*}  

Table \ref{tab:res_bleu} reports the \textsc{Bleu} scores, as
calculated by \newcite{post18call}'s \texttt{SacreBleu} of the
different models we consider, both on canonical and non-canonical test
sets. Contrary to the first results of \newcite{michel2018mtnt}, the
quality of UGC translation does not appear to be so bad: the drop in
performance observed on non-canonical corpora is of the same order of
magnitude as the drop observed when translation models are applied to
out-of-domain data. For instance, the \textsc{Bleu} score of a
\texttt{Transformer} model trained on \texttt{OpenSubtitles} has the
same order of magnitude on \crapbank, \texttt{MTNT} and \texttt{news}:
on all these datasets, the performance dropped by roughly 4 \textsc{Bleu}
points compared to its performance on in-domain data. However, this
improvement in translation quality is partly due to our \unk
replacement strategy, a step that was not considered by
\newcite{michel2018mtnt}.

As expected, all models perform better when they are trained on the
\opensubs corpus than when they are trained on
\wmt, as the former is intuitively more similar to UGC data
than \texttt{WMT} is. More surprisingly, it appears that
character-based models are largely outperformed by BPE-based models
for most train-test combinations and, therefore, that their capacity to
learn word representations that are robust to noise can be questioned.

Another interesting observation is that, while the
\texttt{Transformer} achieves the best results on all test sets when
trained on \texttt{OpenSubtitles}, it is outperformed by
\texttt{seq2seq} when the \wmt training set is considered. The
situation is similar for character-based models.  This observation
suggests that these models do not capture the same kind of information
and do not have the same generalization capacities, as they roughly
have the same number of parameters (69M parameters for the
\texttt{char2char} and \texttt{Transformer} models, 65M for the
\texttt{charCNN} and 49M for the \texttt{seq2seq} model).

\begin{figure*}[!htpb]
  \centering
  
   \begin{tikzpicture}[scale=.9]

     \begin{groupplot}[ 
        group style={
          group size=2 by 1,
        },
        ]
        \nextgroupplot[
        title={(a) Worst translation hypotheses},
        draw=white,
        ybar=0pt,
        bar width=4pt,
        enlargelimits=0.15,
        ylabel={Counts},
        x tick label style={font=\tiny,rotate=75,anchor=east,},
        symbolic x coords={char del/add,{missing diacritics},phonetic writing,tokenization,wrong tense,special char,agreement,casing,emoji,named entity,contraction,repetition},
        xtick=data,
        ] {
          \addplot coordinates {(char del/add,147) 
            (missing diacritics,82) 
            (phonetic writing,123) 
            (tokenization,104) 
            (wrong tense,48) 
            (special char,164) 
            (agreement,26) 
            (casing,204) 
            (emoji,16) 
            (named entity,236) 
            (contraction,67) 
            (repetition,52)};
          \addplot coordinates {(char del/add,122)
            (missing diacritics,112)
            (phonetic writing,106) 
            (tokenization,102)
            (wrong tense,61)
            (special char,176)
            (agreement,14)
            (casing,132)
            (emoji,25)
            (named entity,207)
            (contraction,66)
            (repetition,65)};
        }
        \coordinate (c1) at (rel axis cs:0,1);
        \nextgroupplot[
        title={(b) Best translation hypotheses},
        draw=white,
        ybar=0pt,
        bar width=4pt,
        enlargelimits=0.15,
        yticklabel=\empty,
        legend style={
          at={($(0,0)+(1cm,1cm)$)},
          legend columns=2,
          fill=none,draw=black,
          anchor=center,
          align=center},
        legend to name=fred,
        x tick label style={font=\tiny,rotate=75,anchor=east,},
        symbolic x coords={char del/add,{missing diacritics},phonetic writing,tokenization,wrong tense,special char,agreement,casing,emoji,named entity,contraction,repetition},
        xtick=data,
        ] {
          \addplot coordinates {(char del/add,29)
            (missing diacritics,32)
            (phonetic writing,16)
            (tokenization,20)
            (wrong tense,13)
            (special char,48)
            (agreement,12)
            (casing,9)
            (emoji,10)
            (named entity,180)
            (contraction,29)
            (repetition,23)};
          \addlegendentry{\texttt{\small char2char}}
          \addplot coordinates {(char del/add,26)
            (missing diacritics,27)
            (phonetic writing,33)
            (tokenization,22)
            (wrong tense,24)
            (special char,73)
            (agreement,8)
            (casing,13)
            (emoji,5)
            (named entity,166)
            (contraction,22)
            (repetition,11)};
          \addlegendentry{\texttt{\small Transformer}}
        }c
        \coordinate (c2) at (rel axis cs:1,0);
      \end{groupplot}
      \coordinate (c3) at ($(c1)!.5!(c2)$);
      \node[below] at (c3 |- current bounding box.south) {\pgfplotslegendfromname{fred}};
    \end{tikzpicture}
  \caption{Comparison of the number of UGC specificities in the best
    and worst translation hypotheses of \texttt{char2char} and
    \texttt{Transformer}.  Noise categories are defined in
    Table~\ref{tab:error_types}. \label{fig:errors_stats}}
\end{figure*}

\paragraph*{Error analysis \label{sec:analysis}} In order to find which kind of UGC
specificities are the most difficult to translate and can explain the
difference in performance between character-based and BPE-based
systems, we have conducted a contrastive analysis between the
predictions of the \texttt{Transformer} and the \texttt{char2char}
models. For each model, we have selected the 100 source sentences with
the highest translation quality and the 100 ones with the lowest
translation quality.\footnote{To evaluate translation quality at the
  sentence level, we consider the edit distance between the
  translation hypothesis and the reference.} We have
manually annotated these 400 sentences, using the typology described
in Table~\ref{tab:error_types}, to identify which UGC specificities
were the hardest to translate. \review{2}{4- ok- ds}{Such a classification is adopted 
from the fined-grained list of UGC particularities of \newcite{sanguinettietal:2020:treebankingUGC}.}
Examples of annotations are given in
Table~\ref{tab:output_examples}, in the Appendix.

\begin{table}
  \footnotesize
  \begin{tabular}{ll}
    \toprule
    code & kind of specificities \\
    \midrule
    1 & Letter deletion/addition\\
    2 & Missing diacritics \\
    3 & Phonetic writing\\
    4 & Tokenisation error\\
    5 & Wrong verb tense\\
    6 & \#; @, URL\\
    7 & Wrong gender/grammatical number\\
    8 & Inconsistent casing\\
    9 & Emoji\\
    10 & Named Entity\\
    11 & Contraction\\
    12 & Graphemic/punctuation stretching\\
    \bottomrule
  \end{tabular}
  \centering
  \caption{Typology of UGC specificities used in our manual
    annotation.\label{tab:error_types}}
    \vspace{-1em}
\end{table}

Figure~\ref{fig:errors_stats} shows the number of UGC specificities in
the 100 worst and 100 best translations of the two considered
models. As expected, there are far fewer specificities in the best
translations than in the worst translations except for the high number
of named entities that is found in both cases. This is, however, not a
surprise given the high number of named entities in UGC (see
\textsection\ref{sec:model}).

\begin{table*}[h!]
{\footnotesize

  \begin{tabularx}{\textwidth}{llX}
    \toprule
    \ding{195} & src              & Si au bout de 5 mois \textbf{tu rescend \blue{(3)} toujours se \blue{(3)} genre} "d'effet secondaire" c'est vraiment mauvais.\\
      & ref              & If after 5 months you're still feeling this kind of "side effect" that's really bad. \\
      & Tx               & \underline{If after} five \underline{months you're still} re-exciting \underline{this} whole "special \underline{effect}" thing is \underline{really bad}.\\
      & c2c              & In the end of five \underline{months}, \underline{you're} always responding to secondary effects, \underline{that's  really bad}.\\
      \midrule   
          \ding{196} & src & y a ma cousine qui \textbf{joue a \blue{(2)}} \textbf{flappy bird \blue{(10)}} \textbf{mdrrrrrrrrrrr \blue{(11, 12)} elle et plus nuuul \blue{(12,7)} que moi}\\
      & ref              & my cousin plays flappy bird loooooooooool she's more hopeless than me \\
      & Tx               & There's \underline{my cousin} who \underline{plays} a \underline{flappy bird} mdrrrrrrrrrrrrrrrrrrrrrrrrrrrrrrrrrrrrrrrrrrrrrrrrrrrrrrrrrrrr\\
      & c2c              & There's \underline{my cousin} that \underline{plays} a \underline{flappy bird} flappy bird might bring her and nouse \underline{than me}.\\
      \bottomrule    
  \end{tabularx}
}
  \caption{Examples from our noisy UGC corpus showing the \transformer
    (denoted as Tx) and \chartchar (denoted as c2c)
    predictions. Source sentences have been annotated with UGC
    specificities of Table 1 
    (in blue). Part of
    the reference that were correctly generated are underlined.
    \label{tab:output_examples}}
\end{table*}


The specificities that are the most problematic to
translate\footnote{Because we only report statistics on the best and
  worst translations of each system, our analysis does not allow us to
  directly compare the performance of the two models and we can only
  identify general trends in the impact of UGC idiosyncrasies on
  translation quality.} appear, as can be seen in
Figure~\ref{fig:errors_stats}.a, to be the presence of named-entities (\nth{10}
category) and the inconsistent casing (\nth{8} category) often
corresponding to several words written fully upper-cased to denote
emotions or excitement. Counterintuitively, these two kinds of noise
have more impact on the \texttt{char2char} model than on the
\texttt{Transformer} model, even if it could be expected that the
character embeddings learned by the former would not be sensitive to
the case. When manually inspecting its prediction, it appears that,
often, the \texttt{char2char} model is not able to copy named entities
and replace some of their characters. For instance, the entity
\textit{Cllaude468} is translated into \textit{Claudea64666}, and
\textit{flappy bird} into \textit{flappe bird}. Interestingly this
second entity is correctly translated by the \texttt{char2char} model
when it is written \textit{Flappy Bird}. Similarly, the
\texttt{Transformer} model often tries to translate part of the
hashtags relying on the subword units identified by the BPE
tokenization rather than simply copying them from the source: for
instance ``\textit{\textbf{\#Ça}MeVenereQuand}'' 
(``\textbf{\#It}AnnoysMeWhen'' in English) is translated into 
``\textit{\#CaMevenre\textbf{When}}''\draftremove{or
``\textit{\textbf{\# It's} true}'', and ``\textit{\#TT}'' into 
``\textit{\#TTT}'' or ''\textit{\#T.T}}.

Another important category of errors, is the \nth{6} category that
corresponds to hashtags, mentions and URLs, for which the
\texttt{char2char} model is not capable of producing characters or
sequence of characters that are rare in the training set (namely
\texttt{\#}, \texttt{@} or \texttt{http://www}). For instance, the
\texttt{char2char} model only outputs 8 sharp symbols when translating
the test set of the \crapbank, while the reference contains 345 hash
tags starting with a~`\#'. While the \texttt{Transformer} model is
less sensitive to this problem (it produces 105 sharp symbols when
translating the \crapbank test), its translation quality also suffers
from the presence of hashtags as, as for named entities it often
translates some of the subword units resulting from the BPE
tokenization.  Finally, we notice that errors 2 and 12
(diacritization and repetition) are treated somewhat better by the
\texttt{char2char} model than by the \texttt{Transformer} model, being
less frequent in the worst translations of the
former. 

\paragraph*{Qualitative Analysis}

Table~\ref{tab:output_examples} reports translation hypotheses
predicted by the \texttt{Transformer} and \texttt{char2char} models.
These examples illustrate the difficulty of the \texttt{char2char}
model to translate named entities: while the simple strategy of
copying unknown OOVs from the source implemented by the
\texttt{Transformer} is very effective, the \texttt{char2char} model
tend to scramble or add letters to NE. They also illustrate the impact
of phonetic writing on translation: for instance ``\textit{rescend}''
that should be spelt ``\textit{resents}'' ({\em feeling}, example 4) and
``\textit{joue a} that should be spelt ``\textit{joue à}'' ({\em play a} vs
{\em play with} - auxiliary verb instead of a preposition- example 5)
both result in a wrong translation: in the former case, the meaning of
the sentence is not captured, in the latter the ``\textit{a}'' is
wrongly copied in the translation hypothesis.

\subsection{Copy Task Control Experiment }

To corroborate our analysis of the special characters impact on
\texttt{char2char} and quantify the impact of \texttt{char-OOVs} and of
rare character sequences, we conduct a control experiment in which a
\texttt{char2char} system with different vocabulary sizes must learn to
copy the source, that is to say: we apply a \texttt{char2char} on an
artificial parallel corpus in which the source and the target
sentences are identical. By varying the number of characters
considered by the system, we can control the number of rare characters
in our corpora (recall that with a char-vocabulary of size $N$ all
characters but the $N$ most frequent ones are mapped to a special \unk
character). Note that this task is a simplification of the problem we
have highlighted in the previous section: the model has simply to
learn to always make an exact copy of the whole source sentence and
not to detect some special characters (such as \# or @) from the input 
that trigger the copy of the next characters, while the rest of the
sentence should be translated.

More precisely, we built an artificial train set containing 1M random
sentences with lengths between 5 and 15 characters long, keeping a 164
fixed-size character vocabulary (this correspond to the size of the extended
ASCII alphabet), \review{1}{2  -ok ds}{and whose characters are distributed uniformly,
 in order to rule out the impact of rare characters and keeping only 
 the effect of \texttt{char-OOVs} over the performance}. 
 We consider two test sets made of 3,000 sentences
each: \texttt{in-test} that uses the same 164 characters as the train
set and \texttt{out-test} that uses 705 different characters.  Source
and reference are identical for every example of the train and test
sets.

\paragraph{Results} Table~\ref{tab:test_copy_res} reports the results achieved on the
copy task with and without replacing the predicted \unk
symbols. 

\iffalse
Reducing the vocabulary size results in an increase of the
\textsc{Bleu} score on the two considered conditions, even before
performing \unk replacing. \hl{These observations suggest that unknown or
rare characters are not only difficult to copy, but they also distort
the representation of the input sequence built by the system impacting
the whole prediction.} 
\else 
Note that, in this very simple task, \unk
characters are always replaced by their true value.  These results
show that this task is not trivial for char-based systems: even when
all characters have been observed during training, the system is not
able to copy the input perfectly.

Above all, reducing the vocabulary size from 164 to 125 results
in an increase of the \textsc{Bleu} score on the two considered
conditions, even without replacing the \unk that the system has
started to generate, \review{2}{5 - ok ds}{where `\%\unk{} pred.' indicates 
the percentage of \unk tokens in the prediction}. Further reducing the size of the vocabulary
artificially improves the quality of the systems: they generate more
and more \unk, which are replaced during post-processing by their true
value. These observations suggest that unknown or rare characters
  are not only difficult to copy, but they also distort the
  representation of the input sequence built by the system impacting
  the whole prediction.  

\fi

\begin{table}[h!]
\centering
{\footnotesize
\begin{tabular}{lrrrrr}
  \toprule
  & \multicolumn{5}{c}{Vocabulary Size} \\
  \cline{2-6}
                      & 164  & 125   & 100  & 80   & 60 \\
  \midrule
  \texttt{in-test} \\
  \phantom{ab} \%\unk{} pred.  & 0    & 0.2  & 5    &   17 & 29.5 \\
  \phantom{ab} \textsc{Bleu}   & 92.9 & 95.8 & 77.6 & 24.9 & 1.9 \\
  \phantom{ab} +\unk{} rep.    & 92.9 & 96.6 & 98.4 & 98.5 & 98.7 \\
  \midrule
  \texttt{out-test} \\
  \phantom{ab} \%\unk{} pred.  & 0     & 9.2  & 13.8 & 25   & 36 \\
  \phantom{ab} \textsc{Bleu}   & 54.5  & 63.7 & 52.3 & 15.3 & 0.9 \\
  \phantom{ab} +\unk{} rep.    & 54.4  & 96.6 & 98.7 & 99.1 & 99.5 \\
  \bottomrule
\end{tabular}
}
\caption{Results of the copy task evaluated by the \textsc{Bleu} score
  before and after \unk replacement ({\em +\unk{} rep.}) and percentage of \unk{} characters in the prediction ({\em \%\unk{} pred.}).   \label{tab:test_copy_res}}
\end{table}

\section{Impact of the Char-Vocabulary Size on Translation Quality \label{sec:cv}}
\begin{figure}[htbp]
  \centering
  \begin{tikzpicture}[scale=.9]
    \begin{axis}[
      draw=white,
      ybar,
      legend style={
        at={(0.5,1.05)},
        anchor=south,
        legend columns=2},
      enlargelimits=0.15,
      ylabel={\textsc{Bleu} score},
      xlabel={\# \texttt{charOOVs}},
      symbolic x coords={0,1,$\geq 2$},
      xtick=data,
      nodes near coords,
      every node near coord/.append style={rotate=90, anchor=west},
      ]
      \addplot coordinates {(0,24.8) (1,19.8) ($\geq 2$,16.9)};
      \addlegendentry{\texttt{\small char2char}}
      \addplot coordinates {(0, 25.8) (1, 29.7) ($\geq 2$, 30.2)};
      \addlegendentry{\texttt{\small Transformer}}
    \end{axis}
  \end{tikzpicture}
  \caption{Impact of the number of \texttt{char-OOVs} in the input
    sentence on the \textsc{Bleu} score. Systems trained on \opensubs.\label{fig:score_by_coov}}
\vspace{-0.6em}
\end{figure}
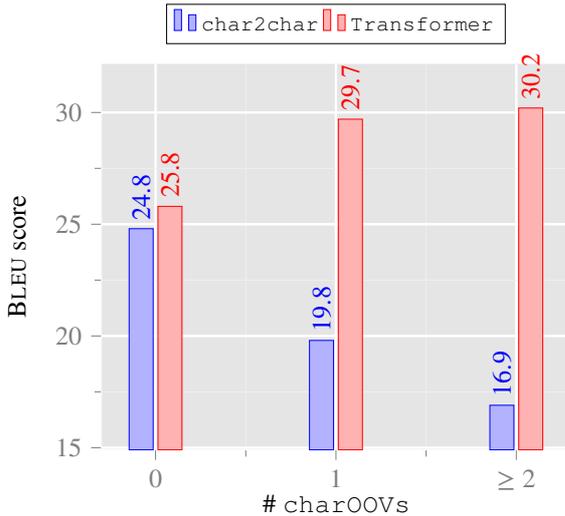

To validate in a more realistic setting the conclusion we draw from
our copy tasks experiments, we present in
Figure~\ref{fig:score_by_coov}, the impact of unknown characters on
translation quality of the \crapbank test set for the
\texttt{Transformer} and \texttt{char2char} models. While the
translation quality is almost the same for both models when there is
no \texttt{char-OOVs} in the input, the occurrence of an unknown
character hurts the translation quality of the \chartchar model beyond
the mistranslation of a single character.

Motivated by these observations, we carried out experiments on our
\texttt{char2char} models, trained on \texttt{OpenSubtitles}, with
different character vocabulary size.  Results in
Table~\ref{tab:results_vocab} show that `correctly' choosing the
vocabulary size can improve the translation performance on both
canonical and non-canonical corpora.  We did not observe any
noticeable improvements in in-domain data as \draftreplace{our vocabulary size hyper-parameter
is always much smaller than the number of different characters
observed on our two training corpora (cf. column \texttt{\#chars} of
Table~\ref{tab:stat_corpus})}{those test sets contain very few
\texttt{char-OOVs} compared to their UGC counterparts, {\em e.g.,} 0.76\% for
the \texttt{OpenSubtitles} test set.} However, overall stability in
in-domain performance is observed, unlike what has been reported when
 domain-adaptation methods were used \cite{michel2018mtnt,DBLP:conf/iclr/BelinkovB18}.

 For vocabulary sizes larger than 90
characters, the \texttt{char2char} system achieves the same
\textsc{Bleu} scores as with 90 characters.  Note that in our first
experiments (reported in Table~\ref{tab:res_bleu}), we used the
`standard' character vocabulary size (namely 300 characters) that was
used in~\cite{DBLP:journals/tacl/LeeCH17_et_al} and, to the best of
our knowledge, in all following works.

\begin{table}[!htpb]
  \centering
  {\footnotesize
    \begin{adjustbox}{max width=\columnwidth}
    \begin{tabular}{rrrrr}
      \toprule
      vocab. size & \crapbank & \texttt{MTNT} & \texttt{News} & \texttt{OpenTest$^\dag$} \\
      \midrule
      90 & 23.9& 25.8& 18.7 &  26.6\\
      85 & 23.9 & 25.3& \textbf{19.9}&  \textbf{26.9}\\
      80 & 23.9 & 25.8& 18.3& 26.6 \\
      75 & 24.5 & \textbf{25.9} & 17.8&  26.3\\
      70 & \textbf{24.6} & 25.4 & 17.8&  26.3\\
      65 & 22.7 & 25.5 & 18.0&  26.4\\
      \bottomrule
    \end{tabular}
    \end{adjustbox}
    \caption{\textsc{Bleu} results for MT of \texttt{char2char} with reduced
      vocabulary size. Systems trained on \opensubs. {\em $^\dag$ marks
      in-domain test set}.\label{tab:results_vocab}}
  }
\end{table}

As shown by an evaluation done on the blind tests of our UGC dataset, in Table~\ref{tab:results_blind}, setting the vocabulary
size parameters based on the improvement they bring reduces the difference in
performance between the \chartchar and \seqtseq systems. With the `correct' vocabulary size the
former is outperformed by only 0.3 (resp. 1) \textsc{Bleu} points
while, with the default value, the difference is 1.2 (resp. 3.1)
\textsc{Bleu} points on the \crapbank (resp. \texttt{MTNT}).

\begin{table}
\centering
{\footnotesize
   \begin{tabular}{rrr}

     & \makecell{\crapbank\\blind} & \makecell{MTNT\\blind} \\
     \midrule
     \texttt{char2char-base} &     17.8     &     20.9   \\
     \phantom{abc}\texttt{vocab-75} & 18.3 & 24.0 \\
     \phantom{abc}\texttt{vocab-70} & 18.7 & 22.8 \\
\midrule
     \texttt{\textit{\gray{Transformer}}} & \textit{\gray{ 19.0}} & \textit{\gray{25.0}}  \\
     \texttt{\textit{\gray{seq2seq}}} & \textit{\gray{22.1}} & \textit{\gray{20.4}} \\

     \bottomrule
   \end{tabular}
   \caption{\textsc{Bleu} results for blind MT test sets of \chartchar
     with reduced vocabulary size. Systems trained on \opensubs. \gray{{\em The \transformer
     and \seqtseq are presented as comparison data points.}}\label{tab:results_blind}}
         \vspace{-1.5em}
}
\end{table}

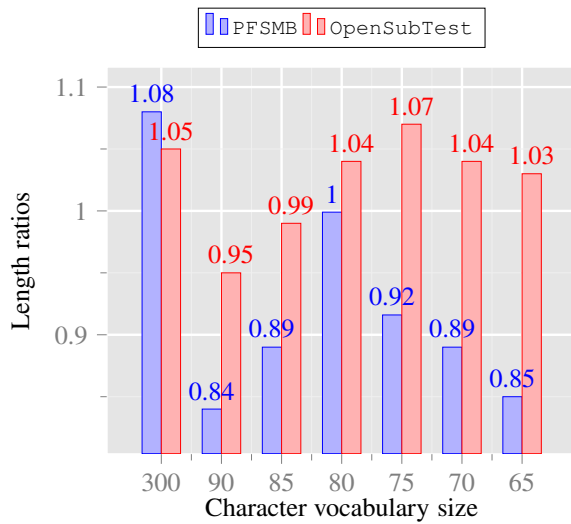
\begin{figure}[htpb]
\centering
\begin{tikzpicture}[scale=.9, fontscale/.style = {font=\small}]
  \begin{axis}[
    draw=white,
    ybar=0pt,
    bar width=8pt,
    enlargelimits=0.15,
    legend style={
      at={(0.5,1.05)},
      anchor=south,
      legend columns=2},
    ylabel={Length ratios},
    xlabel={Character vocabulary size},
    symbolic x coords={300, 90, 85, 80, 75, 70, 65},
    xtick=data,
    nodes near coords,
    nodes near coords align={vertical},
    ]
    \addplot coordinates {(300,1.08) (90,0.84) (85,0.89) (80,0.999) (75,0.916) (70,0.89) (65,0.85)};
    \addlegendentry{\texttt{\small \crapbank}}
    \addplot coordinates {(300,1.05) (90,0.95) (85,0.99) (80,1.04) (75,1.07) (70,1.04) (65,1.03)};
    \addlegendentry{\texttt{\small OpenSubTest}}
\end{axis}
\end{tikzpicture}
\caption{Reference/hypothesis length ratios on the \crapbank test set
  produced by \chartchar for different vocabulary
  sizes. \label{fig:length_ratio}}
\end{figure}

By manually comparing the system outputs for different vocabulary
sizes, it seems that the observed differences in performance can be
explained by drop- and over-translation phenomena, two well-identified
limits of NMT~\cite{DBLP:conf/emnlp/MiSWI16,DBLP:conf/aclnmt/KoehnK17,
  DBLP:conf/ijcnlp/LeMYM17}. An analysis of
Figure~\ref{fig:length_ratio}, which displays the ratio between the
hypothesis and reference lengths for the different vocabulary sizes,
seems to confirm this hypothesis as it appears that the vocabulary
size parameter provides control over the translation hypothesis length
and consequently, a way to limit these drop- and over-translation
phenomena.


\section{Discussion}
\label{sec:disc}

Our work follows many studies on the robustness of character-based models, most notably  from \newcite{DBLP:conf/iclr/BelinkovB18} who compared different translation unit levels for NMT and showed that, overall, MT systems trained on
clean data only presented considerable performance decrease when
processing synthetic character-level noise and word-level noise.  Here, we also perform comparisons with other mainstream strong BPE-based baselines and because of our annotated UGC data set are able to provide a more precise view on the phenomena at stake when processing {\em natural} noisy-input.

Similarly to \newcite{DBLP:conf/naacl/DurraniDSBN19}  what concluded for
morpho-syntactic classification tasks, our results show that a more fined-grained granularity of learning representations, characters over BPEs, provides a higher robustness to certain types of noise.
Contrary to the both aforementioned works, our study is performed
using annotated  real-world noisy UGC, which proved crucial for our study.

In line to \newcite{DBLP:conf/coling/FujiiMAHMSI20} findings, where fine-grained NMT granularity provided robustness advantages when processing misspelling, our results show that the best and worst translations' specificities distribution point to a better performance of char2char for the \textit{missing diacritics} category, giving insights on more specific types of misspellings that affect performance.

Continuing this research track, we broaden the spectrum of studying UGC specificities, exploring the effects of the vocabulary training size and show that tuning it can achieve better results when translating noisy UGC. This simple hyper-parameter choice  proved effective, providing an alternative to using fine-tuning
methods \cite{michel2018mtnt} or back-translation \cite{DBLP:conf/naacl/VaibhavSSN19}.



\section{Conclusion}

\iftrue
We showed that in zero-shot scenarios, char-based models are not robust to UGC idiosyncrasies. We presented several experiments that explained this counter-intuitive
result by an over-sensibility of these models to the vocabulary size.
We demonstrated that drastically lowering this parameter increased the
robustness of our char-based models when facing noisy UGC while
maintaining almost the same level of performance for in-domain and out-of-domain canonical datasets. Interestingly, we noticed a lack in the literature regarding the importance of vocabulary considerations when training character-based NMT models. However, our results show that a path for improvement exists, leading to more generic and stable models and making us believe that further research on the matter can bring promising alternatives to domain-adaptation via target domain data addition. Our data set is available at \url{https://github.com/josecar25/char_based_NMT-noisy_UGC}.

\else
In this work, we have tested the capacity of character-based NMT
system to translate user-generated content. We show that, contrary to
what could be expected, character-level models are outperformed by a
`standard' \transformer{} model. The detailed comparison of the
translations of a character-based and a \transformer{} models even
show that the former is very sensitive to UGC idiosyncrasies and is
not able to translate texts that are productive in terms of new forms,
new structures or new domains, even if it was specifically designed as
an open-vocabulary model.

It must, however, be noted that all the results presented in this work
have been achieved in the difficult, but realistic, zero-shot scenario
in which all models are trained on canonical data only. We plan, in
our future work, to find ways to automatically augment training data
to make models less sensitive to UGC idiosyncrasies.
\fi

\section*{Acknowledgments}
We thank the reviewers for their valuable feedback. This work was funded by the French Research Agency via the ANR ParSiTi project (ANR-16-CE33-0021).

\bibliographystyle{acl_natbib}
\bibliography{ugc_at_charlevel}

\clearpage
\onecolumn
\appendix
\section{Appendix}

\begin{table*}[h!]
{\scriptsize

  \begin{tabularx}{\textwidth}{llX}
    \toprule
    \texttt{\crapbank}& src       & ma \textbf{TL} se vide après \textbf{j'me fais chier} puis \textbf{jme} sens seule mais \textbf{c} surtout \textbf{pdt les vacs mais c pas le cas dc} ça compte pas trop\\ 
      & ref              & my TL is emptying then I get bored then I feel alone but mostly during holidays but it's not the case so it's not so important \\
      & google               & my TL is empty after I'm pissing then I feel alone but c especially pdt vacs but it does not matter that it does not count too much \\ \cmidrule{2-3}
      &src & Dans Flappy Bird quand mon Bird il fait 10 jsuis trop contente \#TeamNul Mdddr \\
      &ref & At Flappy Bird when my Bird has 10 Im so happy \#TeamFail Loool \\
      &google & In Flappy Bird when my Bird is 10, I'm too happy \#TeamNul Mdddr \\ \cmidrule{2-3}
      
      &src& \textbf{Boooooooooooooooooooooonnne Annniversaure Ma viiiiiiiiiiiiiie jtm} plus que tout profite bien de tes \textbf{22ans moaaaaaaaaa} \\
      &ref & Haaaaaaaaaaaaaaaaaaaaaapppy Biiirthday My liiiiiiiiiiiiiife luv you more than anything enjoy your 22years mwaaaaaaaaah \\
      &google & Boooooooooooooooooooooonnne Annniversaure My viiiiiiiiiiiiiie jtm more than anything enjoy your 22 years moaaaaaaaaaa  \\

    \midrule
    \texttt{MTNT} & src        & \textbf{AJA} que le XV de France féminin est\textbf{ un train de faire} un grand \textbf{chelem, OKLM}\\  
      & ref          & TIL that the XV of Feminine France is doing a grand chelem, FRESH\\
      & google              & AJA that the XV of France feminine is a train of a grand slam, OKLM \\ \cmidrule{2-3}
      
      &src &   Je pensais mais c'est le même \textbf{ident et mdp} que mon compte "normal", et il détecte même la \textbf{profile pic} et le nom \\
      &ref &  I thought so but it's the same username and password as my 'normal' account, and it detects even the profile pic and the name \\
      &google &  I thought but it's the same ident and mdp as my "normal" account, and it even detects the pic profile and the name \\ \cmidrule{2-3}
      
      &src&  \textbf{Aaaaaaaah.... 8 ans après, je viens de percuter....  :o 'tai} je me disais bien que je passais à côté d'un truc vu les \textbf{upvotes.} \\
      &ref &  Aaaaaaaah.... 8 years later, I've just realized.... :o damn I had the feeling that I was missing something considering the upvotes. \\
      &google & Aaaaaaaah .... 8 years later, I just collided ....: oh well I was telling myself that I was missing something seen the upvotes. \\

      \bottomrule    
  \end{tabularx}
}
  \caption{Examples from both noisy UGC showing the source phrase, reference translation and Google Translate output. UGC idiosyncrasies are highlighted using bold characters. 
    \label{tab:UGC_examples}}
\end{table*}

\begin{table*}[h!]
{\footnotesize

  \begin{tabularx}{\textwidth}{llX}
    \toprule
\iftrue
    \ding{192} & src       & \textbf{JohnDoe389 \blue{(10)}} qui n'arrive pas \textbf{a \blue{(2)} depasser \blue{(2)}} 1 a \textbf{\blue{(2)}} \textbf{FlappyBird \blue{10}} ... \textbf{ptddddr \blue{(11, 12)}}\\ 
      & ref              & JohnDoe389 who can't score more than 1 at FlappyBird ... lmaooooo \\
      & Tx               & John Doe389 \underline{who can't} pass one  to Flappy Bird... \underline{ptdddr.} \\
      & c2c              & Johndeigh doesn't happen to pass one on Flappyrib... \underline{please.} \\
 
    \midrule
    \ding{193} & src        & \textbf{JohnDoe665 \blue{(10)} viens \blue{(5)}} de regarder \textbf{Teen Wolf (2011) S03E17 [Silverfinger] \blue{(10)}} et s'en va apprendre l'humanit\'e \`a \textbf{Castiel \blue{(10)}}. \\  
      & ref          & JohnDoe665 just watched Teen Wolf (2011) S03E17 [Silverfinger] and he's on his way to teach Castiel humanity.\\
      & Tx           & John Doe665 \underline{just watched Teen Wolf (2011) S03E17} (\underline{Silverfinger}) \underline{and} is going \underline{to teach humanity} at \underline{Castiel}. \\
      & c2c          & Johndoedsoids is looking at \underline{Teen Wolf 2011}, and learn about \underline{humanity} in \underline{Castiel}. \\ 
      \midrule
\ding{194} & src              & \textbf{Jai \blue{(1,4)} fait 10 a \blue{(2)}} \textbf{flappy bird \blue{(10)} mddr \blue{(11, 12)}} \textbf{\# JeMeLaDonneMaisJavancePas \blue{{(6)}}} \\
      & ref              & I did 10 at flappy bird lool \# JeMeLaDonneMaisJavancePas \\
      & Tx               & \underline{I did 10} a \underline{flappy bird} mdr \# I'm not moving\\
      & c2c              & \textmusicalnote I've made  \underline{10} \underline{flappy} birdd \textmusicalnote \\
      \midrule
\else
    \ding{195} & src              & Si au bout de 5 mois \textbf{tu rescend \blue{(3)} toujours se \blue{(3)} genre} "d'effet secondaire" c'est vraiment mauvais.\\
      & ref              & If after 5 months you're still feeling this kind of "side effect" that's really bad. \\
      & Tx               & \underline{If after} five \underline{months you're still} re-exciting \underline{this} whole "special \underline{effect}" thing is \underline{really bad}.\\
      & c2c              & In the end of five \underline{months}, \underline{you're} always responding to secondary effects, \underline{that's  really bad}.\\
      \midrule   
          \ding{196} & src & y a ma cousine qui \textbf{joue a \blue{(2)}} \textbf{flappy bird \blue{(10)}} \textbf{mdrrrrrrrrrrr \blue{(11, 12)} elle et plus nuuul \blue{(12,7)} que moi}\\
      & ref              & my cousin plays flappy bird loooooooooool she's more hopeless than me \\
      & Tx               & There's \underline{my cousin} who \underline{plays} a \underline{flappy bird} mdrrrrrrrrrrrrrrrrrrrrrrrrrrrrrrrrrrrrrrrrrrrrrrrrrrrrrrrrrrrr\\
      & c2c              & There's \underline{my cousin} that
      \underline{plays} a \underline{flappy bird} flappy bird might
      bring her and nouse \underline{than me}.\\
\fi
      \bottomrule    
  \end{tabularx}
}
  \caption{Examples from our noisy UGC corpus showing the \transformer
    (denoted as Tx) and \chartchar (denoted as c2c)
    predictions. Source sentences have been annotated with UGC
    specificities of Table 1 
    (in blue). Part of
    the reference that were correctly generated are underlined.
    \label{tab:output_examples}}
\end{table*}

\pagebreak

\section{Reproducibility}
\paragraph*{Data}
All the UGC test sets and source code for our experiments are provided in the supplementary materials. For training data, we let the reader refer to each project's website for \texttt{WMT}\footnote{\url{https://www.statmt.org/wmt15/translation-task.html}} (consisting of \texttt{Europarlv7}, \texttt{Newcommentariesv10} and \texttt{Open Subtitles}\footnote{\url{http://opus.nlpl.eu/download.php?f=OpenSubtitles/v2018/moses/en-fr.txt.zip}}, both accessed on November, 2019. Regarding clean test sets, we used \texttt{newstest15} from \texttt{WMT} and a subset of 11.000 unique phrases from \texttt{Open Subtitles}. We make the former test available in the link provided above for exact performance comparison.

\paragraph*{Computation}
The NMT systems were trained using 1 Tesla V100, during an average of 72 hours to converge to the final solution for the \texttt{char2char} model and 56 hours for the BPE-based baselines. 
\subsection{NMT Models}
\paragraph*{Character-based models}

Both \texttt{charCNN} and \texttt{char2char} models were trained as
out-of-the box systems using the implementations provided by
\cite{DBLP:conf/aaai/KimJSR16} for
\texttt{charCNN},\footnote{\url{https://github.com/harvardnlp/seq2seq-attn}}
an by \cite{DBLP:journals/tacl/LeeCH17_et_al} for
\texttt{char2char}.\footnote{\url{https://github.com/nyu-dl/dl4mt-c2c}}

\paragraph*{BPE-based models}

We consider, as baseline, two standard NMT models that take, as
input, tokenized sentences. The first one is a \texttt{seq2seq}
bi-LSTM architecture with global attention decoding. The
\texttt{seq2seq} model was trained using the XNMT
toolkit~\cite{DBLP:conf/amta/NeubigSWFMPQSAG18}.\footnote{We decided
  to use XNMT, instead of OpenNMT in our experiments in order to
  compare our results to the ones of \newcite{michel2018mtnt}.} It
consists of a 2-layered Bi-LSTM layers encoder and 2-layered Bi-LSTM
decoder. It considers, as input, word embeddings of 512~components and
each LSTM units has 1,024 components. 

We also study a vanilla Transformer model using the implementation
proposed in the OpenNMT framework
\cite{DBLP:conf/amta/KleinKDNSR18}. It consists of 6 layers with word
embeddings of 512 components, a feed-forward layers made of 2,048
units and 8 self-attention heads.

The vocabulary parameter for experiments in \textsection~\ref{sec:cv} were obtained through a first exploratory uniform segmentation of the possible vocabulary sizes, and then, discovering the threshold (vocabulary with 90 characters) for which the \texttt{char2char} model started producing \unk during evaluation. We then proceeded to obtain the results displayed above by sampling this parameter by decreasing 5 characters.

\paragraph*{Hyper-parameters}
In Table~\ref{tab:hyperparams}, we list the training variables set for our experiments. They match their corresponding default hyper-parameters.
\begin{table}[!htpb]
\centering
\begin{tabular}{l|l}
\toprule
   Batch size  & 64 \\
    Optimizer & Adam \\
    Learning rate & 1e-4 \\
    Epochs    &   10 (best of) \\
    Patience   &  2 epochs \\
    Gradient clip & [-1.0, 1.0] \\
    \bottomrule
\end{tabular}
\caption{Hyper-parameters used for training the NMT systems. \label{tab:hyperparams}}
\end{table}


\paragraph*{Pre-processing}
For the BPE models, we used a 16K merging operations tokenization employing \texttt{sentencepiece}\footnote{\url{https://github.com/google/sentencepiece}}. For word-level statistics we segmented the corpora using the \texttt{Moses} tokenizer \footnote{\url{https://github.com/moses-smt/mosesdecoder}}.

\end{document}